\documentclass[letterpaper, 10 pt, conference]{ieeeconf}  

\IEEEoverridecommandlockouts                              

\usepackage{graphics} 
\usepackage{epsfig} 
\usepackage{mathptmx} 
\usepackage{times} 
\usepackage{amsmath} 
\usepackage{amssymb}  
\usepackage{tikz}
\usepackage{tikz-3dplot}
\usetikzlibrary{arrows.meta, calc, decorations.markings}
\usepackage{xcolor}
\usepackage{hyperref}
\usepackage{svg}
\usepackage{caption}
\usepackage{subcaption}
\usepackage{booktabs}
\usepackage{cite}
\title{\LARGE \bf
{\bf GM3:} A General Physical Model for Micro-Mobility Vehicles
}

\author{Grace Cai$^{*1}$, Nithin Parepally$^{*1}$, Laura Zheng$^{1}$, and Ming C. Lin$^{1}$
\thanks{$^{1}$The authors are with the Department of Computer
Science, University of Maryland at College Park, MD,
U.S.A.
        Email: \{glcai, nparepa\}@terpmail.umd.edu, \{lyzheng, lin\}@umd.edu} %
\thanks{$^{*}$ Equal contribution}%
}

\begin{document}

\maketitle
\thispagestyle{empty}
\pagestyle{empty}

\begin{abstract}
    Modeling the dynamics of micro-mobility vehicles (MMV) is becoming increasingly important for training autonomous vehicle systems and building urban traffic simulations. However, mainstream tools rely on variants of the Kinematic Bicycle Model (KBM) \cite{mathworksBicycleKinematicModel, autowareMpcLateralController,li2024tactics2d, highwayEnv} or mode-specific physics that miss tire slip, load transfer, and rider/vehicle lean. To our knowledge, no unified, physics-based model captures these dynamics across the {\em full} range of common MMVs and wheel layouts. We propose the ``Generalized Micro-mobility Model'' (GM3), a tire-level formulation based on the tire brush representation~\cite{svendenius2003brush, deur20043d, svendenius2004brush} that supports arbitrary wheel configurations, including single/double track and multi-wheel platforms. We introduce an interactive model-agnostic evaluation and visualization framework that decouples vehicle/layout specification from dynamics to compare the GM3 with the KBM and other models, consisting of fixed step RK4 integration, human-in-the-loop and scripted control, real-time trajectory traces, and logging for analysis. We also empirically validate the GM3 on the Stanford Drone Dataset's deathCircle (roundabout) scene~\cite{robicquet2017learning} for biker, skater, and cart classes.
\end{abstract}

\section{INTRODUCTION}
Micro-mobility vehicles (MMVs) such as bicycles and scooters are becoming increasingly central to today's urban traffic. In 2023, people took 133 million trips on shared micro-mobility across the US and 24 million trips in Canada, led by the rise of dockless and station-based shared systems~\cite{NACTOSharedMicromobility2023}. Beyond bikes and e-scooters, adoption of other forms of micro-mobility, including skateboards and low-speed vehicles (LSVs), is also on the rise~\cite{LSPTVaug2024, OIAStateOutdoorMarket2022}. 

At the same time, accidents involving micro-mobility and motor vehicles are a growing safety concern. According to the National Highway Traffic Safety Administration (NHTSA), in 2023, an estimated 49,989 U.S pedal cyclists were injured with 1,166 fatalities, 81\% of which occurred in urban areas~\cite{NHTSA2025bicyclists}. For powered micro-mobility specifically, the U.S. Consumer Product Safety Commission (CPSC) reported an estimated 448,600 emergency department (ED) visits associated with e-bikes, e-scooters, and hoverboards between 2017 and 2023, with motor-vehicle collisions cited as the leading cause of fatalities. E-bike-related ED visits alone increased tenfold over this period~\cite{tark2023micromobility}). 

From a modeling perspective, the USDOT Intelligent Transportation Systems, Joint Program Office found that current Analysis, Modeling, and Simulation (AMS) tools are largely vehicle-centric, with limited support for multi-modal interactions and inadequate representation of pedestrians, bicylists, and other micro-mobility modes~\cite{ali2024complete}. The International Transport Forum (ITF) reported that, in California, deployments of autonomous vehicles, 85\% of manual disengagements involving public-space users (AV-VRU interactions) were caused by the vehicle's incorrect perception of user intent~\cite{botero2024lost}. These gaps indicate that existing simulators and traffic models are not suited to capture the complexity of interactions between MMV and autonomous systems well. As a result, this is limiting their use for training and validating autonomous systems, as well as limiting the ability to make MMVs {\em autonomous} too!

Furthermore, many state-of-the-art systems default to some variant of the Kinematic Bicycle Model (KBM)~\cite{mathworksBicycleKinematicModel, autowareMpcLateralController,li2024tactics2d, highwayEnv} due to its simplicity and low computational cost, performing well under low-speed, low-curvature conditions. However, the KBM yields unrealistic trajectories when MMVs perform aggressive or high-curvature maneuvers, such as sharp turns, hard braking, and rapid lane changes. The model does not capture tire slip, load transfer, or rider lean, and its single-track geometry is not generalizable to other forms of micro-mobility with alternate wheel configurations. These limitations make the KBM insufficient for simulating the full range of MMV behaviors that autonomous systems must anticipate in mixed traffic. 

In this work, we address these limitations by introducing the ``Generalized Micro-Mobility Model'' (GM3), a unified, physics-based model using the brush tire model that (1) operates at the tire level, (2) supports arbitrary wheel layouts (single/double track and multi-wheel platforms) and (3) captures slip, load transfer, and rider/vehicle lean. 

The primary motivation for this work is to train and evaluate autonomous systems for design, prototyping, and testing, where \underline{all} types of micro-mobility agents move and interact realistically. Recent efforts only address specific micro-mobility modes such as a vehicle-e-scooter interaction simulator for general intersection and lane-change risk analysis~\cite{simFWVEInteraction2023}, cyclist (CiL), pedestrian (PiL), and automated vehicle (ViL) in-the-loop testbeds for studying vulnerable road user interactions in a shared virtual space~\cite{kaiserUrbanInteraction2024}, and a validated bicycle simulator for reproducing realistic steer, row, yall, and roll dynamics~\cite{haasnoot2024validation}. Additionally, city-scale agent-based micro-mobility simulators tend to focus on high-level path planning rather than tire-level dynamics~\cite{panagiotis2024MMVsim} and widely used simulators such as SUMO approximate MMV dynamics as either "slow vehicles" or "fast pedestrians"~\cite{roosta2023state}. Thus, to our knowledge, no unified, physics-based framework exists that models tire-level dynamics across the full range of micro-mobility vehicles. \\

The main contributions of this work are:
\begin{itemize}
    \item A {\bf Generalized Micro-Mobility Model (GM3)}, a unified, physics-based model for simulating the dynamics of various micro-mobility vehicles. This model operates at the tire level, capturing friction, tire slip, load transfer, and lean, and supports arbitrary wheel configurations.
    \item An {\em interactive evaluation and visualization Framework} for GM3 systems. The framework enables plug-and-play specification of MMV dynamics models, vehicle geometries, and parameters (e.g. vehicle, tire, lean) and provides real-time control and trajectory visualization.
    \item An {\em empirical evaluation of the GM3} against the KBM baseline on real-world bicycle, skateboard, and cart trajectories showing that, relative to a KBM baseline data GM3 reduces Average Displacement Error (ADE) across modes while achieving comparable discrete Fréchet distance (DFD).
\end{itemize}

\section{RELATED WORKS}

{\bf Kinematic Bicycle Model (KBM)} is a simple and computationally inexpensive model that is accurate for low speeds~\cite{kooijman2008experimental}.  However, it has limitations in modeling high-speed and high-curvature scenarios. \cite{polackKBM2017} demonstrated that KBM only yielded consistent results for lateral acceleration $a_y<0.5\mu g$. In an experiment that aimed to improve KBM for higher lateral accelerations ~\cite{matute2019KBM}, the results show vehicle heading error growing significantly with the curvature due to under-steering. This is because KBM does not account for lean or lateral slip. KBM is also not generalizable to other MMVs with different wheel configurations. 

{\bf Using a Tire Model} as a basis, vehicles with varying numbers of wheels in different positions can be simulated by applying the same tire model to each tire and transforming the resulting forces into the vehicle body frame. \cite{szabo2012vehicle} validated the brush model with a Chrysler Voyager, and the experimental trajectories closely matched the corresponding simulation results for low speeds. In \cite{ozerem2019brush}, the brush model and a thermo-physical brush model were validated on a Formula SAE vehicle in extreme scenarios, such as high-speed lane changes. The qualitative and quantitative results show agreement between both models and the experimental data, with an average fitting error of 6.19\% for the basic brush model and 6.09\% for the thermo-physical model. The semi-empirical Magic Formula -- widely used for commercial purposes -- was also evaluated against the data and only provided a 0.46\% improvement over the brush model. These results demonstrate the viability of the brush model for simulating vehicle dynamics at low and high speeds.

Several implementations of the brush model and other tire models have been designed for specific vehicles. Project Chrono~\cite{chrono} provides a Vehicle class with templates for subsystems, including suspension, steering, and braking, and options for three different classes of tire models: rigid, semi-empirical, and finite element. The user can define a vehicle as a list of axles, each consisting of a set of tires with their own parameters. While this library can handle various wheel configurations, its purpose is to model complex military vehicles. Hence, it does not address rider dynamics, and the computational overhead is excessive for MMVs. There is no general model for MMVs that allows the user to customize the wheel layout similarly to Project Chrono, and this paper intends to fill that gap.

\section{PRELIMINARIES} 

To develop a general model that can extend to any micro-mobility vehicle with any wheel configuration, we start from a basic tire model. A relatively compact physics-based tire model is the brush model that represents the tire tread as independent bristles, each of which deflects under slip and friction forces~\cite{PACEJKA201287}. These bristles simulate the elasticity of the combination of carcass, belt, and actual tread elements of the real tire. 

The brush model is derived from calculating the deflection of each bristle and integrating over the contact patch between the tire and the road surface, assuming a parabolic pressure distribution and a flat road surface for simplicity. Tables \ref{tab:params} and \ref{tab:input} include the minimum parameters and variables required to calculate slip forces for a single tire.

\vspace{-5pt}
\begin{table}[h]
\caption{Physical Parameters in the tire brush model.}
\label{tab:params}
\vspace{-5pt}
    \centering
    \begin{tabular}{clc}
    \toprule
    \textbf{Symbol} & \textbf{Description} & \textbf{Units} \\
\midrule
        $\ell$ & Half contact length & m \\
        $R$ & Tire radius & m \\
        $\gamma$ & Camber angle & rad \\
        $\mu$ & Coefficient of friction & -\\
        $c_p$ & Tread element stiffness per unit area & N/m$^2$ \\
    \bottomrule
    \end{tabular}
    \vspace{-15pt}
\end{table}

\begin{table}[h!]
    \caption{Input Variables in the tire brush model.}
    \label{tab:input}
    \vspace{-5pt}
    \centering
    \begin{tabular}{clc}
    \toprule
    \textbf{Symbol} & \textbf{Description} & \textbf{Units} \\
\midrule
        $F_z$ & Normal load & N \\
         $\Omega$ & Wheel angular velocity & rad/s \\
         $\delta$ & Steering angle & rad \\
         $r$ & Vehicle yaw rate & rad/s \\
         $\alpha$ & Slip angle & rad \\   
    \bottomrule
    \end{tabular}
\end{table}
\vspace{-5pt}

The brush model uses a 3D coordinate system with $x$ pointing forward along the tire's trajectory (longitudinal), $y$ pointing right perpendicular to $x$ (lateral), and $z$ pointing up. For simplicity, we assume an isotropic model in which the tire and road properties are the same in all directions ($c_p=c_{px}=c_{py}$, where $c_p$ is the tread stiffness, and $\mu=\mu_x=\mu_y$, where $\mu$ is the friction coefficient), and a flat road surface. We define the theoretical slip vector $\mathbf{\sigma}$ with practical slip quantities $s_x=-\frac{V_{sx}}{V_x}$ and $\tan \alpha = -\frac{V_{sy}}{V_x}$ (where $V_x$ and $V_y$ are the linear longitudinal and lateral velocities, $V_{sx}$ and $V_{sy}$ are the longitudinal and lateral slip velocities, $s_x$ is the longitudinal slip ratio, and $\alpha$ is the slip angle).
\begin{equation}
    \mathbf{\sigma} 
    = -\frac{\mathbf{V_s}}{V_r}
    = \begin{pmatrix}
        \sigma_x \\
        \sigma_y
    \end{pmatrix} \\
    = \begin{pmatrix}
        \frac{s_x}{1+s_x} \\
        \frac{\tan \alpha}{1+s_x}
    \end{pmatrix}
\end{equation}

where $\mathbf{V_s}$ is the slip velocity vector and $V_r=R\Omega$ is the rolling velocity. We introduce spin $\varphi$, which represents the turn slip that occurs when a tire follows a curved path. For a tire with camber angle $\gamma$, radius $R$, turning radius $r_{\text{turn}}$, and reduction factor $\varepsilon_{\gamma}$ (close to 0 for small tires and close to 1 for large tires), the turn slip is
\begin{equation}
    \varphi = -\frac{1}{r_{\text{turn}}} + (1-\varepsilon_\gamma) \frac{\sin(\gamma)}{R}.
\end{equation}

We simplify the equations by introducing a composite tire parameter:
\begin{equation}
    \psi=\frac{2c_{p}\ell^2}{3\mu F_z}
\end{equation}

The resulting {\em longitudinal force $F_x$, lateral force $F_y$, and aligning moment $M_z$} are as follows.
{\footnotesize
\begin{align}
    F_x &= 
    \begin{cases}
        \label{eq:Fx}
       \mu F_z \frac{\sigma_x}{\sigma} [3\psi\sigma-3(\psi\sigma)^2+(\psi \sigma)^3], & \sigma\le \frac{1}{\psi}\\
        \mu F_z \frac{\sigma_x}{\sigma} ,           & \sigma > \frac{1}{\psi}
    \end{cases} \\
    F_y &= 
    \begin{cases}
        \label{eq:Fy}
        \mu F_z [3\psi^*\sigma_{y}-3(\psi^*\sigma_{y})^2+(\psi^* \sigma_{y})^3] + \frac{2}{3}c_p\ell^3\varphi, & |\sigma_{y}|\le \frac{1}{\psi^*}\\
        \mu F_z \text{sgn}(\alpha),           & |\sigma_{y}| > \frac{1}{\psi^*}
    \end{cases} \\
    M_z &= 
    \begin{cases}
        \label{eq:Mz}
        -\mu F_z\ell\psi^*\sigma_{y} [1-3|\psi^*\sigma_{y}|+3|\psi^* \sigma_{y}|^2-|\psi^*\sigma_{y}|^3], & |\sigma_{y}|\le \frac{1}{\psi^*}\\
        0,           & |\sigma_{y}| > \frac{1}{\psi^*}
    \end{cases}
\end{align}
}

The values $\frac{1}{\psi}$ and $\frac{1}{\psi^*}$ represent the boundary between the adhesion region ($\sigma\le \frac{1}{\psi}$), where the bristles deform without sliding, and the sliding region ($\sigma> \frac{1}{\psi}$), where the tire has reached its maximum available friction. Here, sgn is the sign function, $\sigma = \sqrt{\sigma_x^2+\sigma_y^2}$, and $\psi^*$ is the modified tire parameter for camber angle:
\begin{equation}
    \psi^* = \frac{\psi}{1 - \ell\varphi\psi\text{sgn}(\alpha)}
\end{equation}

\begin{figure}[h!]
\centering
 \begin{tikzpicture}[scale=1.1]
    \draw[thick, ->] (0,1.2) -- (1.2,1.6) node[anchor=south west]{$x$};
    \draw[thick, ->] (0,1.2) -- (0,2.8) node[anchor=east]{$z$};
    \begin{scope}[rotate around={-8:(0,1.2)}]
    \draw[thick, fill=gray!30] (0,1.2) ellipse (0.8 and 1.2);
    \draw[thick, fill=gray!60] (0,1.2) ellipse (0.5 and 0.7);
\end{scope}
    \draw[thick] (-1.5,0) -- (1.5,0);
    \draw[->, thick, black!60!green] (0,0) -- (0.8,0.4);
    \node[black!60!green] at (1,0.5) {$F_x$};
    \draw[->, thick, red] (0,0) -- (0.5,-0.5);
    \node[red] at (0.7,-0.5) {$F_y$};
    \draw[->, thick] (0,-0.8) -- (0,0);
    \node[black] at (0.3,-0.8) {$F_z$};
    \draw[->, thick, blue] (0.2,-0.3) arc (0:-200:0.2);
    \node[blue] at (-0.5,-0.3) {$M_z$};
    \draw[dashed, ->] (0,1.2) -- (1,1);
    \node at (1.1,0.9) {$y$};
    \begin{scope}[rotate around={-8:(0,1.2)}]
    \draw[->, thick, black] (0,1.2) -- (1.5,1.5);
    \node[black] at (1.7,1.5) {$V$};
    \draw[dashed] (0,1.2) -- (1.2,1.8);  
    \draw (0.6,1.35) arc (0:15:0.6);
    \node at (0.95,1.53) {$\alpha$};
    \end{scope}
    \draw[->, thick, black] (-0.1,1.3) arc (200:20:0.2);
    \node[black] at (-0.2,1.6) {$\Omega$};
    \draw[dashed] (0,1.2) -- (0,2.4);
    \draw[dashed] (0,1.2) -- (0.2,2.4);
    \draw (0,1.7) arc (90:82:0.5);
    \node at (0.1,2.2) {$\gamma$};
    \draw[thick] (0,1.2) -- (0,0);
    \node[left] at (0,0.8) {$R$};
    \draw[->, very thick] (0,0) -- (0.4,0);
    \node[below] at (0.4,0) {$\ell$};
    \end{tikzpicture}
\caption{A tire modeled with physical parameters and brush model forces and moments.}
\label{fig:tire}
\end{figure}
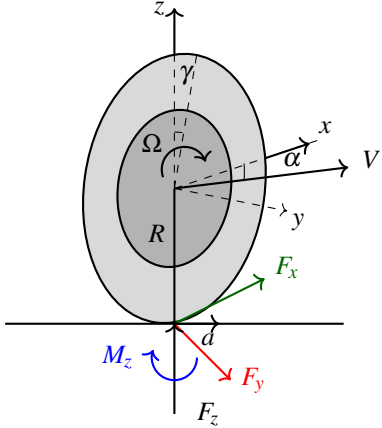

The brush model provides a physics-based approach to predicting tire forces and a balance between computational efficiency and accuracy for vehicle dynamics. Unlike simplified linear tire models that assume constant cornering stiffness, the brush model captures the nonlinear saturation behavior that occurs when tires approach their friction limits -- a critical phenomenon for vehicles that frequently operate in aggressive maneuvering conditions or on varied surface types.

The fundamental insight of the brush model is to represent the tire contact patch as a collection of independent elastic bristles that deflect under slip conditions. This naturally accounts for the transition from elastic deformation at low slip angles to sliding friction at high slip angles, without requiring empirical curve fitting or extensive tire testing data that more complex models demand.

For MMVs, this approach offers three key advantages. First, these vehicles often experience significant slip due to their lightweight nature, small contact patches, and ability to perform aggressive maneuvers. Second, the brush model's relatively simple parameter set makes it practical for real-time control applications while still capturing essential nonlinear effects like force saturation and the coupling between longitudinal and lateral forces. Finally, the model's physical basis allows for intuitive parameter tuning and provides meaningful insights into how tire design choices (contact patch geometry, tread stiffness) affect MMV dynamics.

\section{METHODOLOGY}
In this section, we describe our approach for integrating the brush model and other vehicle dynamics into our Generalized Micro-Mobility Model (GM3). We also include specific dynamics in the cases of single-track MMVs that can lean and skateboards, which have an entirely different steering mechanism.

\begin{figure}[h]
\vspace*{-1em}
  \includegraphics[width=\columnwidth]{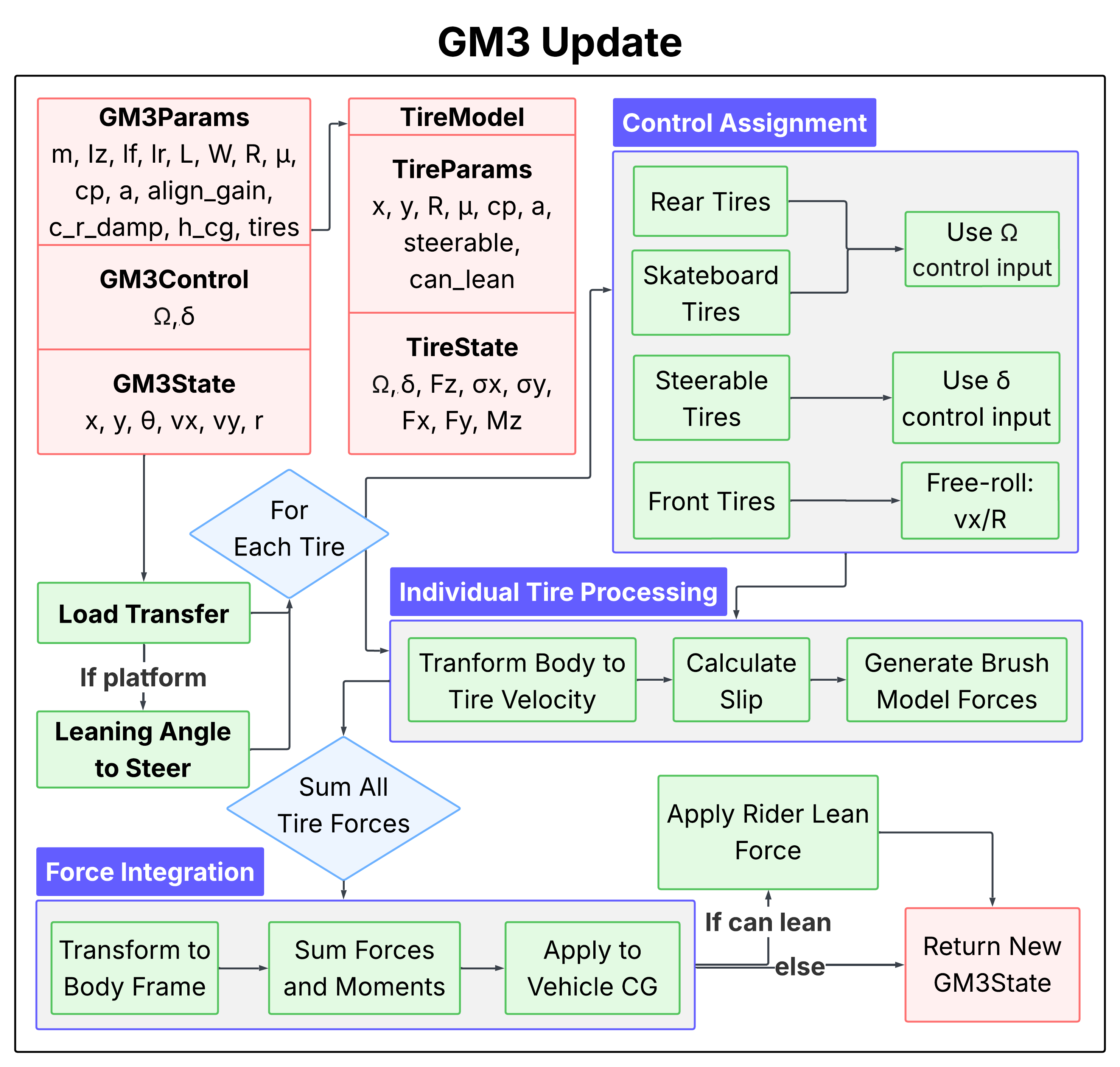}
  \vspace*{-1.5em}
  \caption{The GM3 State gets updated with the following steps: (1) {\bf load transfer} and {\bf leaning angle to steer} conversion for platforms, (2) {\bf control assignment} for each tire, (3) {\bf individual tire processing} with the brush model, and (4) {\bf force integration} and rider lean force application. }
  \label{fig:GM3_update}
  \vspace*{-1.5em}
\end{figure}

\subsection{Vehicle Integration}
When integrating a tire $i$ at position $(x_i,y_i)$ relative to the vehicle center of gravity and steering angle $\delta_i$ into vehicle dynamics, we can transform the vehicle body velocities into tire velocities.
\begin{align}
v_{x,i}^{tire} &= v_{x}^{body} \cos(\delta_i) + (v_{y}^{body} + rx_i) \sin(\delta_i) \\
v_{y,i}^{tire} &= -v_{x}^{body} \sin(\delta_i) + (v_{y}^{body} + rx_i) \cos(\delta_i)
\end{align}

After transforming the body velocities into tire velocities, we use the brush model equations to calculate the forces and moments acting on each tire in its frame. Then, for each tire $i$, the forces in the tire coordinate system are transformed to the vehicle body coordinate system:

\begin{align}
    F_{x,i}^{body} &= F_{x,i}^{tire} \cos(\delta_i) - F_{y,i}^{tire} \sin(\delta_i) \\
    F_{y,i}^{body} &= F_{x,i}^{tire} \sin(\delta_i) + F_{y,i}^{tire} \cos(\delta_i)
\end{align}

$F_{x,i}^{tire}$ and $ F_{y,i}^{tire}$ are the longitudinal and lateral forces in tire $i$'s coordinate system and $F_{x,i}^{body}, F_{y,i}^{body}$ are transformed forces in the vehicle body coordinate system.

The total forces and moments acting on a vehicle with $n$ tires are:

\begin{align}
    F_x^{total} &= \sum_{i=1}^{n} F_{x,i}^{body} = \sum_{i=1}^{n} \left( F_{x,i}^{tire} \cos(\delta_i) - F_{y,i}^{tire} \sin(\delta_i) \right) \\
    F_y^{total} &= \sum_{i=1}^{n} F_{y,i}^{body} = \sum_{i=1}^{n} \left( F_{x,i}^{tire} \sin(\delta_i) + F_{y,i}^{tire} \cos(\delta_i) \right) \\
    M_z^{total} &= \sum_{i=1}^{n} M_{z,i}^{tire} = \sum_{i=1}^{n}(k_{align}M_{z,i}^{tire} + x_i F_{y,i}^{body} - y_i F_{x,i}^{body})
\end{align}

Here, $k_{align}$ is the aligning moment gain factor and $M_{z,i}^{tire}$ is the aligning moment from tire $i$ in the tire coordinate system. We include $k_{align}$ to account for the tire's self-aligning torque acting to turn the wheel back to decrease the slip angle.

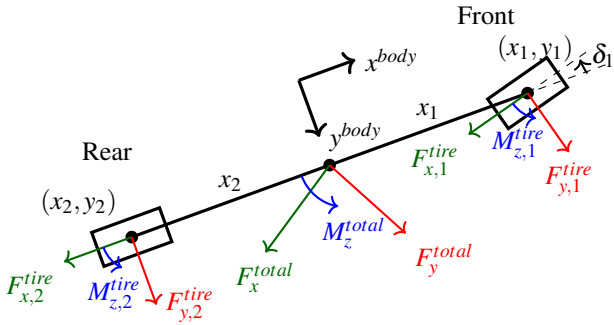
\begin{figure}[h!]
    \centering
    \begin{tikzpicture}[scale=0.8]
        \begin{scope}[rotate around={-160:(0,0)}]
        \draw[very thick] (3,0) -- (6.5,0) node[midway, above] {$x_2$};
        \draw[very thick] (-0.5, 0) -- (3,0) node[midway, above] {$x_1$};
        
        \fill (3,0) circle (3pt);
        
        \draw[dashed] (3,0) -- (-2,0);
        
        \node[below] at (-1,-1) {$(x_1, y_1)$};
        \node[below] at (7,-1.2) {$(x_2, y_2)$};
        
        \draw[very thick, ->] (3,-1.5) -- (2,-1.5) node[right] {$x^{body}$};
        \draw[very thick, ->] (3,-1.5) -- (3,-0.5) node[right] {$y^{body}$};
        
        \begin{scope}[shift={(-0.5,0)}, rotate=15]
            \draw[very thick] (-0.6,-0.3) rectangle (0.6,0.3);
            \fill (0,0) circle (3pt);
            \node[above] at (0,-1.2) {Front};
            \draw[dashed] (0,0) -- (-1.2,0);
            \draw[->, thick] (-1,0.3) arc[start angle=0, end angle=-15, radius=1];
            \node at (-1.4,0.2) {$\delta_1$};
        
            \draw[thick, ->, black!60!green] (0,0) -- (1.2,0) node[below left] {$F_{x,1}^{tire}$};
            \draw[thick, ->, red] (0,0) -- (0,1.2) node[below] {$F_{y,1}^{tire}$};
            \draw[thick, ->,blue] (0.3,0) arc[start angle=0, end angle=45, radius=0.6];
            \node[blue] at (0.6,0.6) {$M_{z,1}^{tire}$};
        \end{scope}
        
        \begin{scope}[shift={(6.5,0)}]
            \draw[very thick] (-0.6,-0.3) rectangle (0.6,0.3);
            \node[above] at (0,-1.2) {Rear};
            \fill (0,0) circle (3pt);
            \draw[thick, ->, black!60!green] (0,0) -- (1.2,0) node[below left] {$F_{x,2}^{tire}$};
            \draw[thick, ->, red] (0,0) -- (0,1.2) node[right] {$F_{y,2}^{tire}$};
            \draw[thick, ->, blue] (0.5,0) arc[start angle=0, end angle=45, radius=0.6];
            \node[blue] at (0.6,0.8) {$M_{z,2}^{tire}$};
        \end{scope}
        
        \draw[thick, ->, black!60!green] (3,0) -- (4.5,1) node[below] {$F_x^{total}$};
        \draw[thick, ->, red] (3,0) -- (2.2,1.5) node[below right] {$F_y^{total}$};
        
        \draw[thick, ->, blue] (3.5,0) arc[start angle=0, end angle=60, radius=0.8];
        \node[blue] at (3,1.2) {$M_z^{total}$};

    \end{scope}
    \end{tikzpicture}
    \caption{Bird's eye view of tire and body forces and moments for a two-wheeled MMV (bicycle or scooter).}
    \label{fig:vehicle}
    \vspace*{-1em}
\end{figure}

\subsection{Load Transfer}
As an MMV accelerates or turns, load transfer occurs between the tires, which must be accounted for to ensure we are passing in the correct normal load $F_z$ to the brush model. Load is transferred to the back wheels when accelerating and to the front wheels when decelerating, maintaining moment equilibrium about the center of gravity. In a double-track vehicle, load is also transferred between the inside and outside wheels when turning. 

For a general MMV, we first redistribute the load longitudinally, then laterally. We start by calculating the static load for the front and rear axles based on their distances to the center of gravity $l_f$ and $l_r$ and the number of tires in the front $n_{f}$ and rear $n_r$. 

\begin{equation}
    F_{z,front} = \frac{mgl_r}{n_{f}L} \qquad 
    F_{z,rear} = \frac{mgl_f}{n_{r}L}
\end{equation}

We use $l_r$ for the front axle and $l_f$ for the rear axle because the axle closer to the center should have more normal force to maintain moment equilibrium.

We calculate the longitudinal load transfer $T_{long}$ and the lateral load transfer using the longitudinal acceleration $a_x$ and the lateral acceleration $a_y$.
\begin{equation}
    T_{long} = \frac{m a_x h_{cg}}{L} \qquad
    T_{lat} = \frac{m a_y h_{cg}}{W}
\end{equation}
    
Here, $h_{cg}$ is the height of the vehicle's center of gravity above the ground, $L$ is the wheelbase (distance between front and rear axles), and $W$ is the track width (distance between left and right wheels on the same axle). $T_{long}$ represents load transferred from front to rear (positive when accelerating forward), and $T_{lat}$ represents load transferred from right to left wheels (positive for rightward acceleration). If there is only one axle, then $T_{long}=0$, and if there is only one tire per axle, then $T_{lat}=0$. The formula to calculate the final load for tire $i$ is as follows.
\begin{equation}
    F_{z,i}= 
    \begin{cases}
        F_{z,front} - \frac{T_{long}}{n_f} - T_{lat}\text{sgn}(y_i) & x_i > 0\\
        F_{z,rear} + \frac{T_{long}}{n_r} - T_{lat}\text{sgn}(y_i) & x_i \le 0
    \end{cases}
\end{equation}

\subsection{Lean}
For MMVs where the rider leans when turning (bicycles, scooters, skateboards, etc.), we must also model the change in position of the rider's center of gravity. We do so by rotating the rider's original center of gravity $(x_{rider,cg}, y_{rider,cg}, h_{rider,cg})$ by lean angle $\phi$ about the x-axis and projecting that into the xy-plane during force calculation. 

The lateral position of the rider's center of gravity when leaning is 
\begin{equation}
    y_{rider,rot} = y_{rider,cg} \cos(\phi) + h_{rider,cg} \sin(\phi)
\end{equation}

The centripetal force acting on the rider during turning is
\begin{equation}
    F_{centripetal} = m_{rider} \cdot a_y
\end{equation}
where $m_{rider}$ is the rider's mass.

Then, the total roll moment about the x-axis contributed by the rider is
\begin{equation}
    M_{x,rider} = y_{rider,rot} m_{rider} g + F_{centripetal} h_{rider,cg}
\end{equation}

\begin{figure}[h!]
    \centering
    \begin{subfigure}[b]{0.28\textwidth}
    \begin{tikzpicture}[scale=1.1]

\draw[very thick] (-1.5,0) -- (1.5,0);

\draw[dashed, ->] (0,0) -- (0,3) node[above] {$z$};
\draw[dashed, ->] (0,0) -- (2,0) node[right] {$y$};
\draw (-0.1,0.1) -- (0.1,-0.1);
\draw (0.1,0.1) -- (-0.1,-0.1);
\node [below right] at (0,0) {$x$};
\node[below left] at (0,0) {$O$};

\fill[gray!30] (0,2.5) circle (4pt);

\coordinate (rider_pos) at ({1.5*sin(25)}, {2.5*cos(25)});
\draw[very thick] (0,0) -- (rider_pos);
\fill (rider_pos) circle (4pt);

\draw[->] (0,1.2) arc[start angle=90, end angle=75, radius=1.2];
\node at (0.2,1.5) {$\phi$};

\draw[dashed] (0,0) -- ({1.5*sin(25)}, 0);
\draw[dashed] (0,2.5) -- ({1.5*sin(25)}, 2.5) node[left, above] {$y_{rider,cg}$};

\draw[very thick, ->] (rider_pos) -- ++(0,-1.5) node[right] {$m_{rider}g$};

\draw[very thick, ->] (rider_pos) -- ++(1.2,0) node[right, below] {$F_{centripetal}$};

\draw[line width=1pt, ->] (0,0.3) arc[start angle=90, end angle=45, radius=0.8];
\node at (1.2,0.2) {$M_{x,rider}$};

\coordinate (y_rot_end) at ({1.5*sin(25)}, 0);
\draw[<->] (0,-0.5) -- (y_rot_end |- 0,-0.5);
\node[below] at ({0.5*1.5*sin(25)}, -0.5) {$y_{rider,rot}$};

\draw[dashed] (0,2.5) -- (-0.5,2.5);
\draw[dashed] ({1.5*sin(25)}, 2.5) -- (-0.5,2.5);
\draw[<->] (-0.4,0) -- (-0.4,2.5);
\node[left] at (-0.4,1.25) {$h_{rider,cg}$};

    \end{tikzpicture}
    \caption{}
    \end{subfigure}%
    \begin{subfigure}[b]{0.2\textwidth}
        \includegraphics[width=\linewidth]{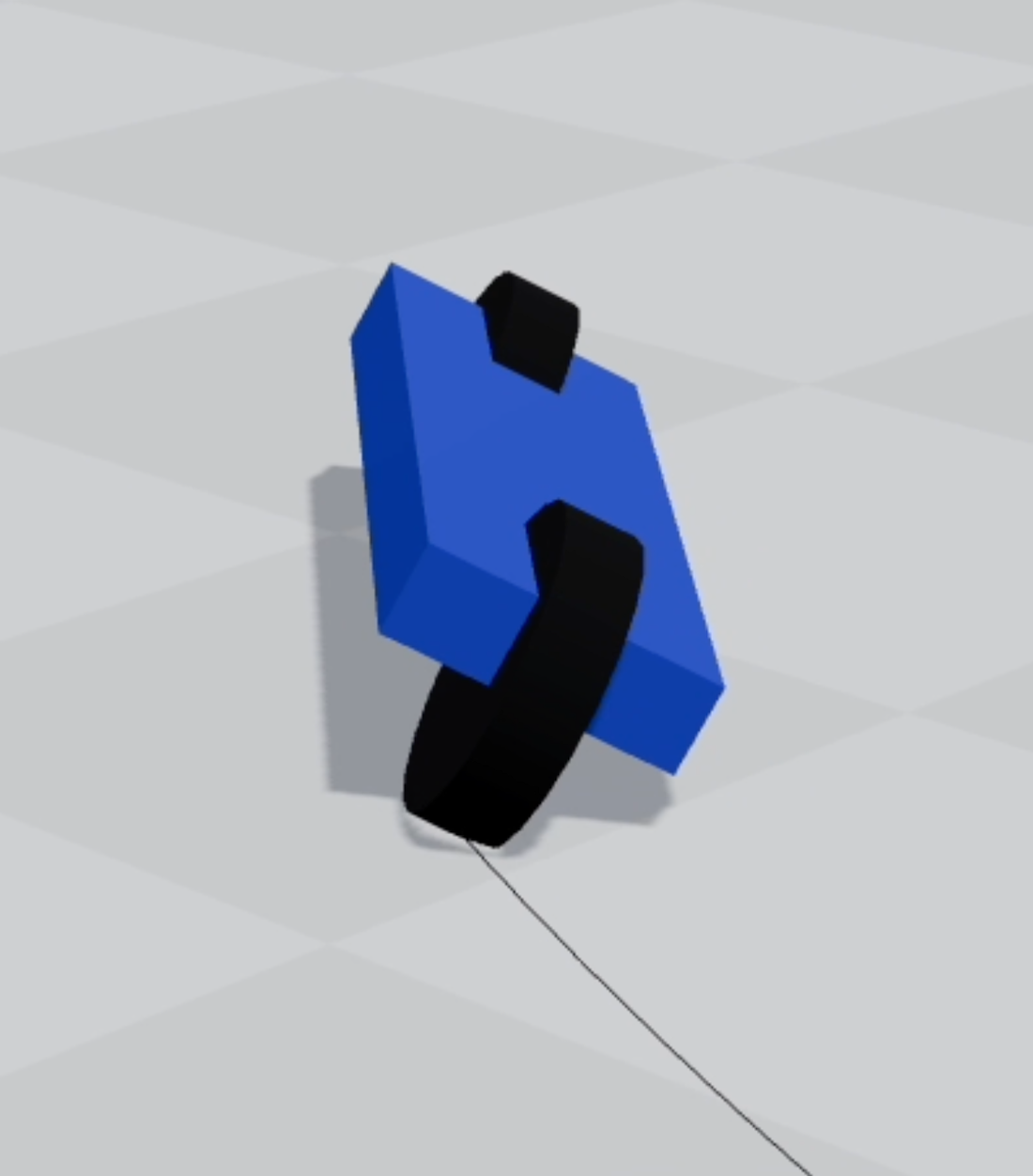}
        \caption{}
    \end{subfigure}%
    \caption{Diagram of rider center of gravity, roll moment, and centripetal force during leaning (a) and visualization with 2-wheel front-to-back layout (b).}
    \vspace*{-1em}
\end{figure}
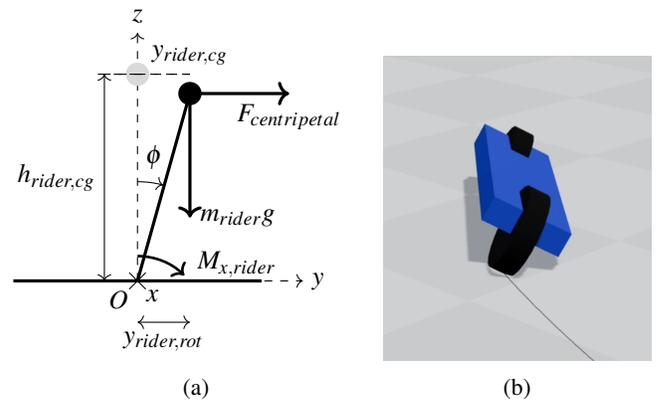

\subsection{Skateboard Truck Geometry}
\label{sec:skateboard}
Skateboards are unique because they use two independent steering assemblies called ``trucks" -- one at each end of the board. Each truck consists of wheels mounted on an axle that can rotate about a kingpin bolt. The rider controls the vehicle's heading entirely through body lean, which causes the trucks to steer in opposite directions.

The kingpin angle $\beta$ is the angle between the kingpin bolt and the vertical plane of the skateboard deck. When the rider leans, this geometry causes the front and rear trucks to steer at equal but opposite angles:
\begin{equation}
    \delta_{f} = k\phi \sin{\beta} \qquad
    \delta_{r} = -k\phi \sin{\beta}
\end{equation}

\begin{figure}[h!]
    \centering
    \begin{tikzpicture}[scale=1]

    \begin{scope}[rotate around={18:(3,2)}] 
    \draw[thick, rounded corners=8pt, fill=brown!20] (0.3,1.2) rectangle (5.7,2.8);
    
    \draw[thick] (1,2) -- (5,2);
    
    \node[below left] at (5.2,2.5) {$A$};
    \draw[thick] (5.5,1.5) -- (4.5,2.5);

    \begin{scope}[rotate around={-45:(5,2)}] 
        \draw[thick, fill=gray!40, opacity=100] (4.2,1.8) rectangle (4.4, 2.2);
        \draw[thick, fill=gray!40, opacity=100] (5.5,1.8) rectangle (5.7, 2.2);
    \end{scope}
    
    \node[below left] at (1.1,2.3) {$B$};
    \draw[thick] (0.5,1.5) -- (1.5,2.5);
    
    \begin{scope}[rotate around={45:(1,2)}] 
        \draw[thick, fill=gray!40, opacity=100] (0.3,1.8) rectangle (0.5, 2.2);
        \draw[thick, fill=gray!40, opacity=100] (1.5,1.8) rectangle (1.7, 2.2);
    \end{scope}
    
    \draw[fill=black] (3,2) circle (0.08);
    \node[above] at (3,2.2) {$C$};
    \node at (3, 1.6) {$l$};
    
    \draw[->, thick, black] (3,2) -- (3.6,2.6);
    \node[above] at (3.8,2.4) {$v$};
    \draw (3.4,2) arc (0:45:0.4);
    \node at (3.6,2.3) {$\delta$};

    \draw[dashed] (5,2) -- (6,2);
    \draw[dashed] (5,2) -- (5.5,2.5);
    \draw (5.5,2) arc (0:45:0.5);
    \node at (5.9,2.2) {$\delta_f$};
    
    \draw (1.5,2) arc (0:-45:0.5);
    \node at (1.7,1.7) {$\delta_r$};
    \draw[->, thick, black] (5,2) -- (5.45,2.45);
    \node[above] at (5.4,2.4) {$v_A$};
    
    \draw[->, thick, black] (1,2) -- (1.5,1.5);
    \node[left] at (1.5,1.3) {$v_B$};
 
    \end{scope}
    
\end{tikzpicture}
    \caption{Skateboard with front and rear steering angles.}
    \label{fig:skateboard}
    \vspace*{-1em}
\end{figure}
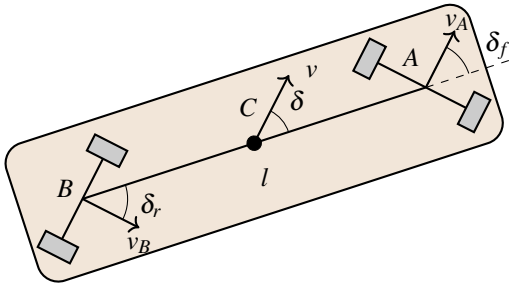

In our simulation, the steering input is used as the lean input to steer the trucks, and all skateboard wheels are driven with the same wheel angular velocity $\Omega$.

\subsection{Generalization to Diverse Vehicle Platforms}

\begin{figure*}
\vspace*{1em}
    \centering
        \begin{subfigure}[b]{0.2\textwidth}
                \includegraphics[width=\linewidth]{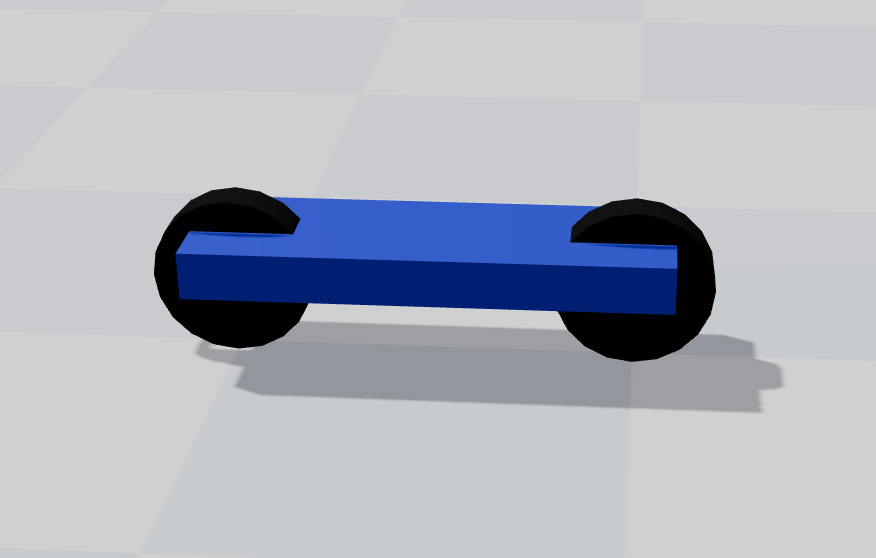}
                \caption{Bicycle (2w)}
                \label{subfig:bicycle}
        \end{subfigure}%
        \begin{subfigure}[b]{0.2\textwidth}
                \includegraphics[width=\linewidth]{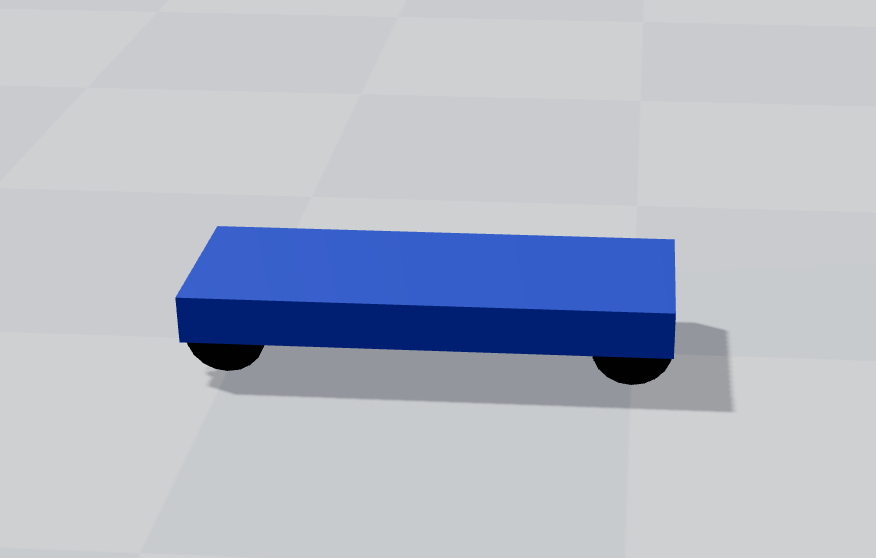}
                \caption{Skateboard}
                \label{subfig:skateboard}
        \end{subfigure}%
        \begin{subfigure}[b]{0.2\textwidth}
                \includegraphics[width=\linewidth]{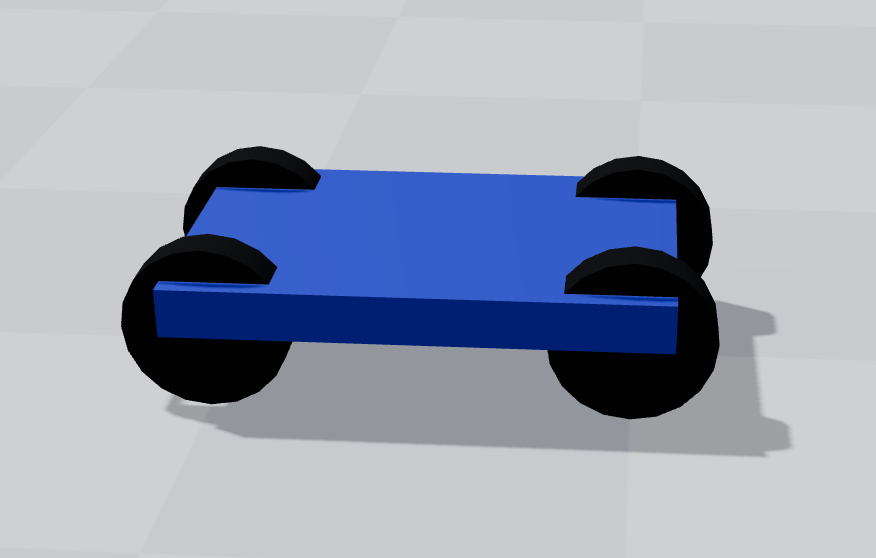}
                \caption{Cart // LSV}
                \label{subfig:cart}
        \end{subfigure}%

        \vspace{1em}

        \begin{subfigure}[b]{0.2\textwidth}
                \includegraphics[width=\linewidth]{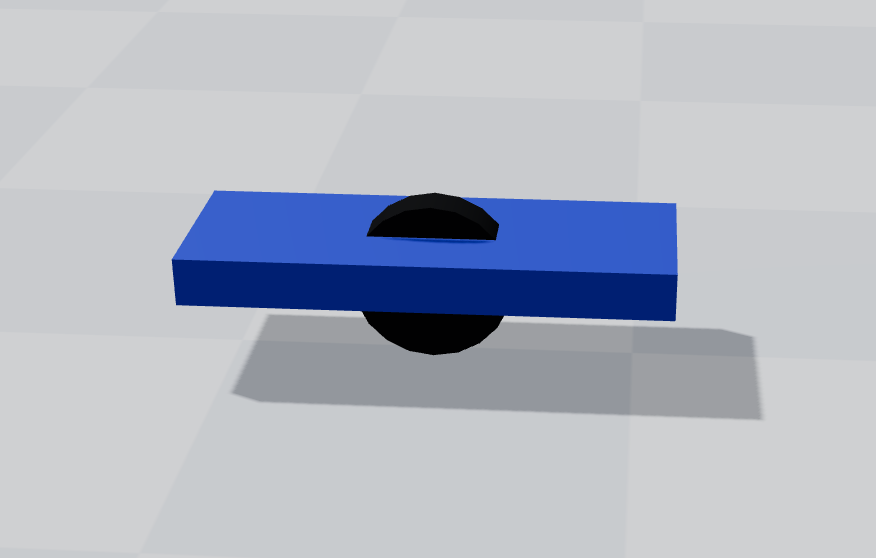}
                \caption{Unicycle}
                \label{subfig:unicycle}
        \end{subfigure}%
        \begin{subfigure}[b]{0.2\textwidth}
                \includegraphics[width=\linewidth]{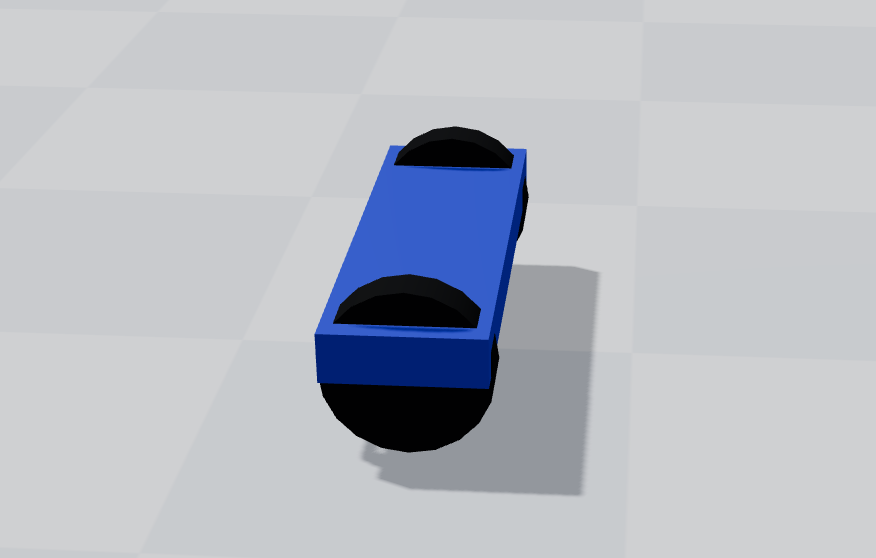}
                \caption{Hoverboard}
                \label{subfig:hoverboard}
        \end{subfigure}%
        \begin{subfigure}[b]{0.2\textwidth}
                \includegraphics[width=\linewidth]{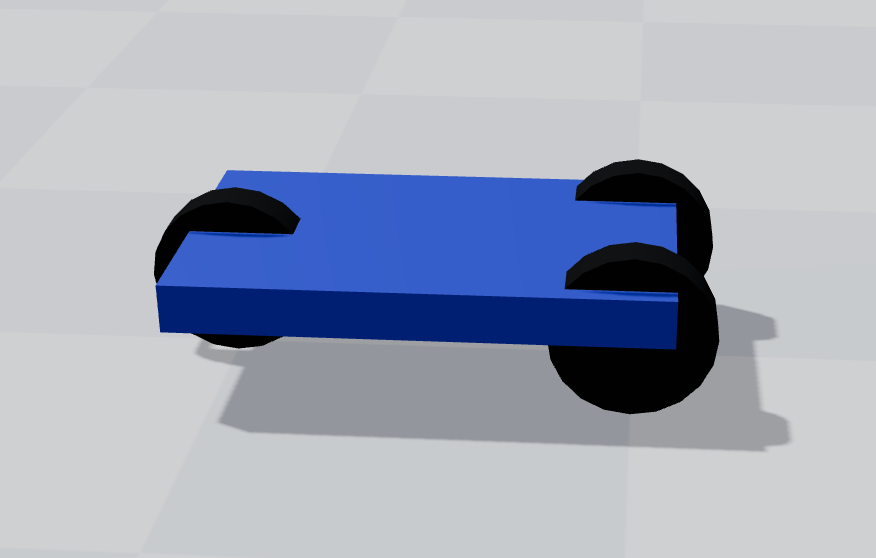}
                \caption{Delta (3w)}
                \label{subfig:delta3}
        \end{subfigure}%
        \begin{subfigure}[b]{0.2\textwidth}
                \includegraphics[width=\linewidth]{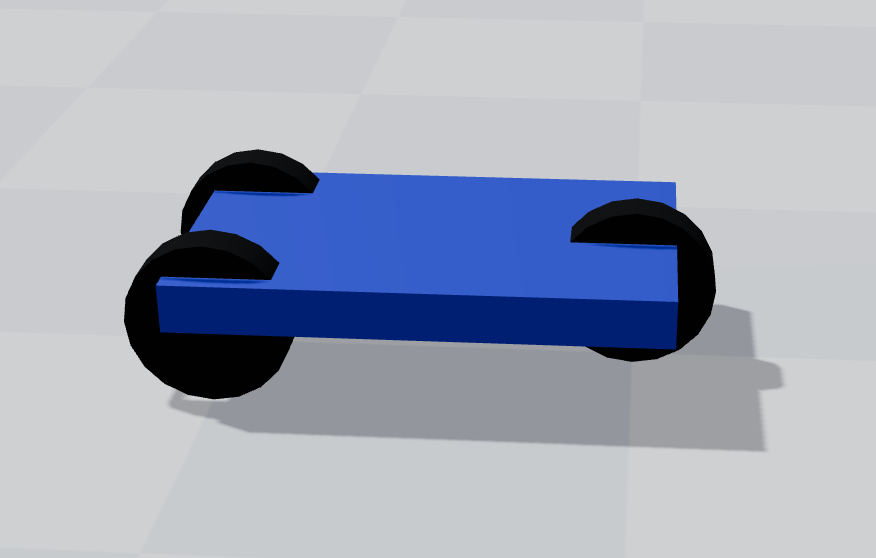}
                \caption{Tadpole (3w)}
                \label{subfig:tadpole3}
        \end{subfigure}%

        \vspace{1em}
        
        \begin{subfigure}[b]{0.2\textwidth}
                \includegraphics[width=\linewidth]{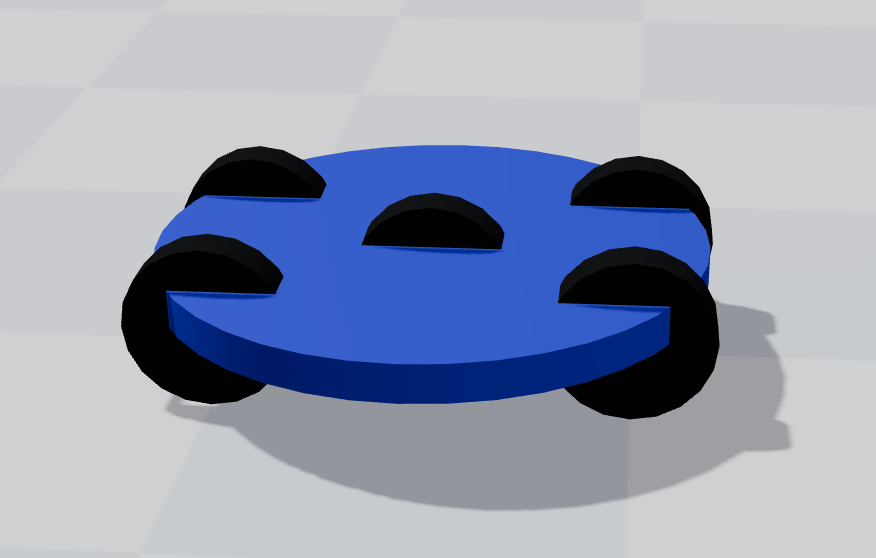}
                \caption{Circular Platform (5w)}
        \end{subfigure}%
        \begin{subfigure}[b]{0.2\textwidth}
                \includegraphics[width=\linewidth]{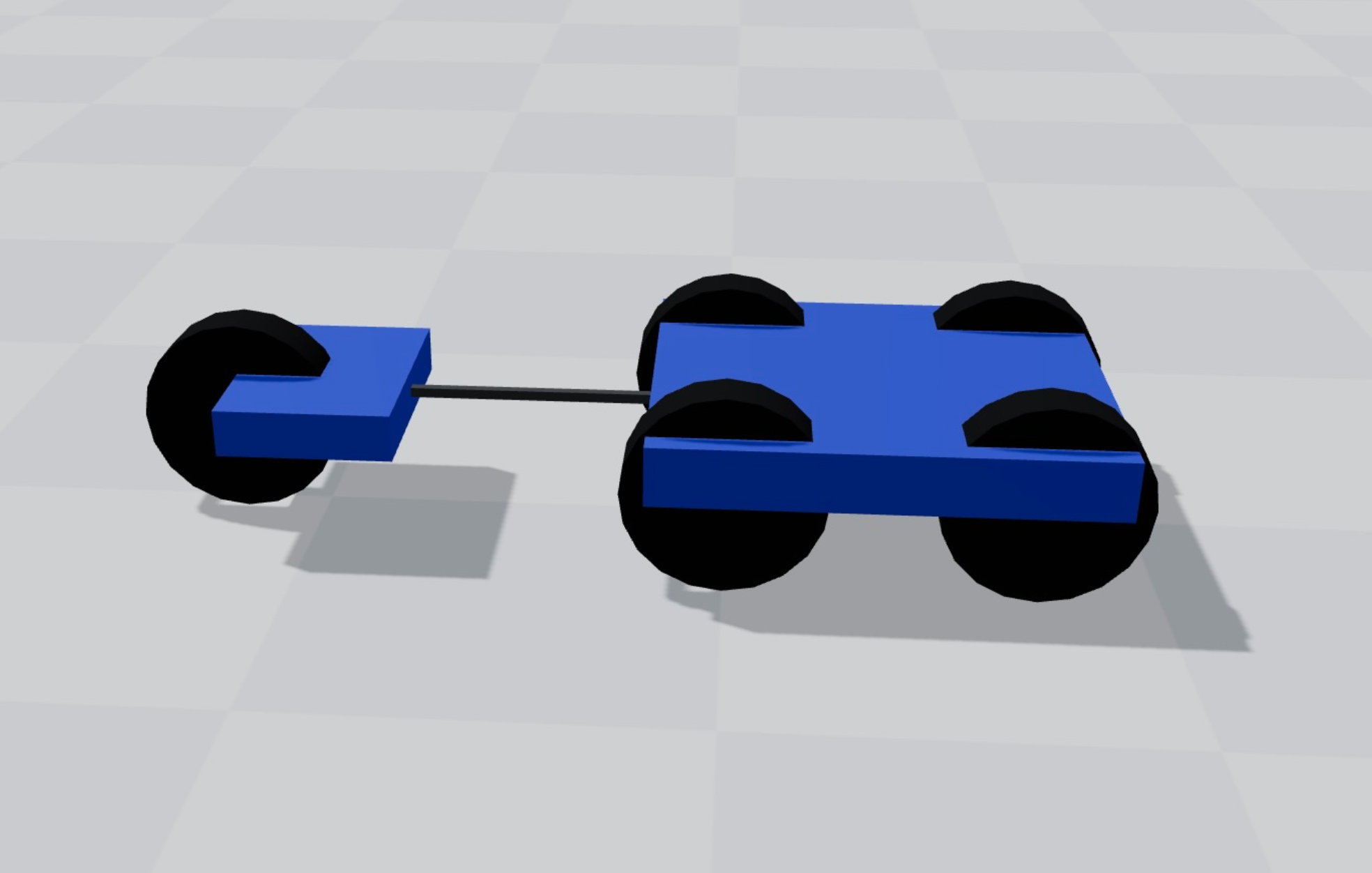}
                \caption{Drawn Carriage}
        \end{subfigure}
        
    \caption{\textbf{Micro-mobility vehicles that can be simulated by the proposed GM3}, illustrating its ability to generalize across arbitrary layouts. Common modes: (a) bicycle/2-wheel scooter, (b) skateboard, (c) cart/LSV.~~Less common: (d) unicycle, (e) hoverboard, (f) delta 3-wheeler, (g) tadpole 3-wheeler. Novel types: (h) 5-wheeled circular platform, (i) drawn carriage.}
    \label{fig:platformDiversity}
    \vspace*{-1em}
\end{figure*}

Our GM3 is widely applicable to model and simulate a diverse range of micro-mobility vehicles across arbitrary layouts. Some common modes are shown in Fig.~\ref{fig:platformDiversity}, including bicycle, scooter, skateboard, cart/LSV, unicycle, hoverboard, delta 3-wheeler, tadpole 3-wheeler, 5-wheeled circular platform, and drawn carriage.

Currently, our model supports three control interfaces: explicit manual control, lean-to-steer with truck geometry, and differential drive. Explicit manual control entails a direct connection between the steering input and the wheels, i.e. turning the handlebars or steering wheel causes the wheels to turn by a proportionate angle. This is the most common control interface and is used for bicycle, scooter, cart, delta 3-wheeler, tadpole 3-wheeler, 5-wheeled circular platform, and drawn carriage. Lean-to-steer control couples lean angle to the steering angle of the wheels as described in section \ref{sec:skateboard}, and is used for skateboard. Lastly, differential drive rotates the wheels at different rates to turn. This control interface is used for hoverboard and can be extended to robots with fixed wheels. 

Since GM3 is modular, the control module can be extended to other control interfaces. This allows GM3 to support any wheeled MMV or robot, given individual wheel steering angles and angular velocity from the control module. 

\subsection{Interactive Evaluation and Visualization Framework}
\begin{figure}[t]
    \centering
    \includegraphics[width=\linewidth]{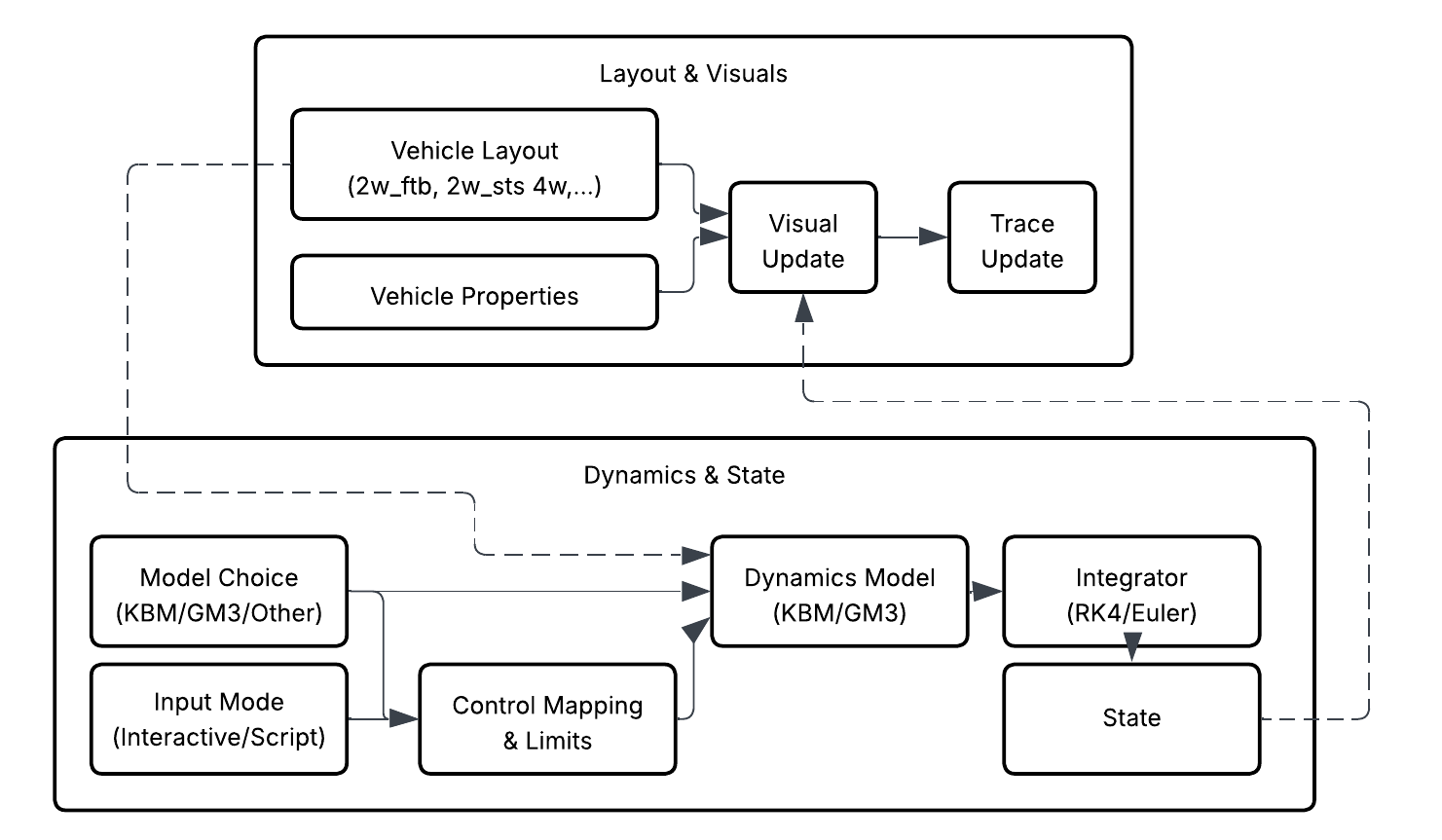}
    \vspace*{-1em}
    \caption{Micro-mobility evaluation and visualization framework diagram.}
    \label{fig:sandbox}
    \vspace*{-1.5em}
\end{figure}

To evaluate the proposed GM3 and facilitate comparison with the KBM baseline, we developed an interactive, model-agnostic evaluation and visualization framework \ref{fig:sandbox}. It is designed to be both a tool for validating the dynamics of different micro-mobility modes and a visualization environment for comparing trajectories produced by different models. Additionally, while this work primarily focuses on comparing GM3 against KBM, the framework is designed to be extensible so additional models can be introduced by adhering to the same interface for control mappings, state, and visual updates.

The framework supports a range of micro-mobility layouts, including single-track vehicles such as bicycles and scooters, side-by-side two-wheel configurations such as hoverboards and segways, three-wheel platforms delta (1F-2R) and tadpole (2F-1R) with front-axle steering, four-wheel carts, and a skateboard configuration, as shown in Fig.~\ref{fig:platformDiversity}.
Steering kinematics are rendered consistently for each layout, with Ackermann steering used for multi-axle vehicles and a skateboard mode that couples front and rear wheel yaw in opposite directions to simulate truck geometry. The idea here is to separate vehicle geometry and model dynamics so that the same platform can be evaluated on multiple MMV models.

Control can be provided in two modes: (1) {\em Interactive} and (2) {\em Script}. In interactive mode, the user can drive the MMV agent using keyboard controls, with steering, speed, and braking intents mapped to model-specific inputs. While steering is consistent across models, longitudinal controls differ by model. KBM accepts acceleration commands while GM3 accepts wheel-speed commands. Input rate limits and saturation are applied for realism (e.g. maximum steer angle, maximum steering rate). In script mode, users define vehicle kind, model type, initial states, parameter values, and control sequences. Scripted runs are intended for qualitative side-by-side comparisons.

At runtime, dynamics are advanced with fixed-step fourth-order Runge-Kutta (RK4) integration. At each time step, inputs or controls are read (from keyboard or a script), transformed into model-specific commands, and fed to the appropriate model to update the state. The new state is then used to append time-aligned state and control values to logs and update the visualization which, beyond vehicle effects, extends an accumulating trajectory trace. The trace can be cleared or reset to rerun maneuvers with different configurations. Run histories can be exported to CSV with model-specific fields for analysis.
\section{EXPERIMENTS}
In our experiments, we evaluate the accuracy and correctness of GM3 in reproducing realistic, physically-ground micro-mobility trajectories, as compared to the KBM. Additionally, we stress test both models across different aggressive maneuvers, including S-curves, U-turns, and left \& right turns.

\textbf{Dataset.}
Experiments are conducted using the Stanford Drone Dataset (SDD) \cite{robicquet2017learning}, a large collection of aerial videos recorded over the Stanford campus. Each video is accompanied by frame-level annotations in the form of axis-aligned bounding boxes and categorical labels for eight classes of road users, including pedestrians, cyclists, skaters, carts, cars, and buses. For this work specifically, we focus on the deathCircle (roundabout) scene since it contains diverse micro-mobility interactions as bicyclists, skateboarders, and carts navigate a shared space alongside pedestrians taking high curvature routes. Image-plane trajectories are converted to metric world coordinates using the scene's pixel-to-meter scale factors \cite{amirian2020opentraj}. For each annotation, we extract the bottom-midpoint of the bounding box, then resample at the native 30 fps \cite{andle2022stanford} to obtain a uniform step size $\Delta t \approx 0.03$s.

Additionally, to demonstrate GM3's performance on high-speed scenarios, we generate a set of trajectories with the following maneuvers: S-curve, left/right turn, and left/right U-turn. The resulting positions were computed using a model-agnostic dynamics model rather than directly integrating the full GM3 or KBM equations. The state is $(x,y,\theta,v)$, with commanded acceleration $a_{\mathrm{cmd}}(t)$ and curvature $\kappa_{\mathrm{cmd}}(t)$ passed through first-order dynamics, $\dot a=(a_{\mathrm{cmd}}-a)/\tau_a$ and $\dot\kappa=(\kappa_{\mathrm{cmd}}-\kappa)/\tau_\kappa$, to produce smooth control response. The constants $\tau_a$ and $\tau_\kappa$ are response times for acceleration and curvature, respectively. Vehicle speed evolves as $\dot v=a-dv$ (with linear drag), heading as $\dot\theta=v\kappa$, and position as $\dot x=v\cos\theta$, $\dot y=v\sin\theta$. To remain physically realistic, curvature is limited by tire-road friction via $|\kappa|\le \mu g / v^2$, so aggressive commands are clipped to feasible lateral acceleration.

\textbf{Vehicle Instantiation.}
Vehicles, one per mode, are instantiated using representative, publicly reported parameters (e.g., wheelbase, track width, tire radius, mass) taken from widely available bicycle, skateboard, and cart specifications on company websites. The bicycle MMV agent uses a single track geometry with lean, the skater uses a skateboard configuration with rider lean, and the cart uses a four-wheel layout and no rider lean. All layouts use load transfer. To prevent overfitting, parameters are fixed per mode and not tuned per track. 

\textbf{Metrics.}
For each trajectory, we quantify model performance with two measures: (1) {\em Average Displacement Error} (ADE) and (2) {\em Discrete Fréchet Distance} (DFD). Average Displacement Error (ADE) is computed as the mean Euclidean Distance between the estimated model positions and the corresponding ground truth positions across all time steps. ADE captures how well the simulated trajectory remains aligned with the true path when both are driven by the same underlying control sequence. To account for overall path similarity, we report discrete Fréchet distance (DFD).

\textbf{Results.}
Table~\ref{tab:results} summarizes ADE and DFD errors across all three modes. GM3 reduces ADE relative to KBM across all modes (Biker: -5.16\%, Skater: -8.77\%, Cart: -6.82\%), which indicates better step-by-step alignment. For DFD, GM3 improves on the Skater (-6.1\%) and Cart (-2.34\%) modes, but is worse on Biker (+11.56\%). Qualitatively, GM3 reproduces better short-term responses to controls due to the tire-level formulation, which aligns with the ADE observations. Since Fréchet distance emphasizes global path similarity, the larger lean angles of cyclists relative to skaters and golf carts introduce lateral biases that accumulate along the extracted bottom-midpoint trajectories, which can inflate the distance. However, aggregated across all modes, GM3 performs only around 1\% worse than KBM with respect to DFD. Furthermore, the campus setting primarily captures low-speed, cooperative interactions on narrow walkways and roadways (even at the roundabout), so hard braking, rapid lane changes, and sustained high-speed, high-curvature turns tend to be rare. As a result, the reported differences likely understate the advantage GM3 provides in higher-speed scenarios with larger slip angles (e.g., bike lanes, downhill stretches, large evasive maneuvers). Please refer to the supplementary video for live side-by-side GM3 and KBM comparisons. 

\begin{figure*}[t]
\centering

\begin{subfigure}{0.32\textwidth}
    \centering
    \includegraphics[width=\linewidth]{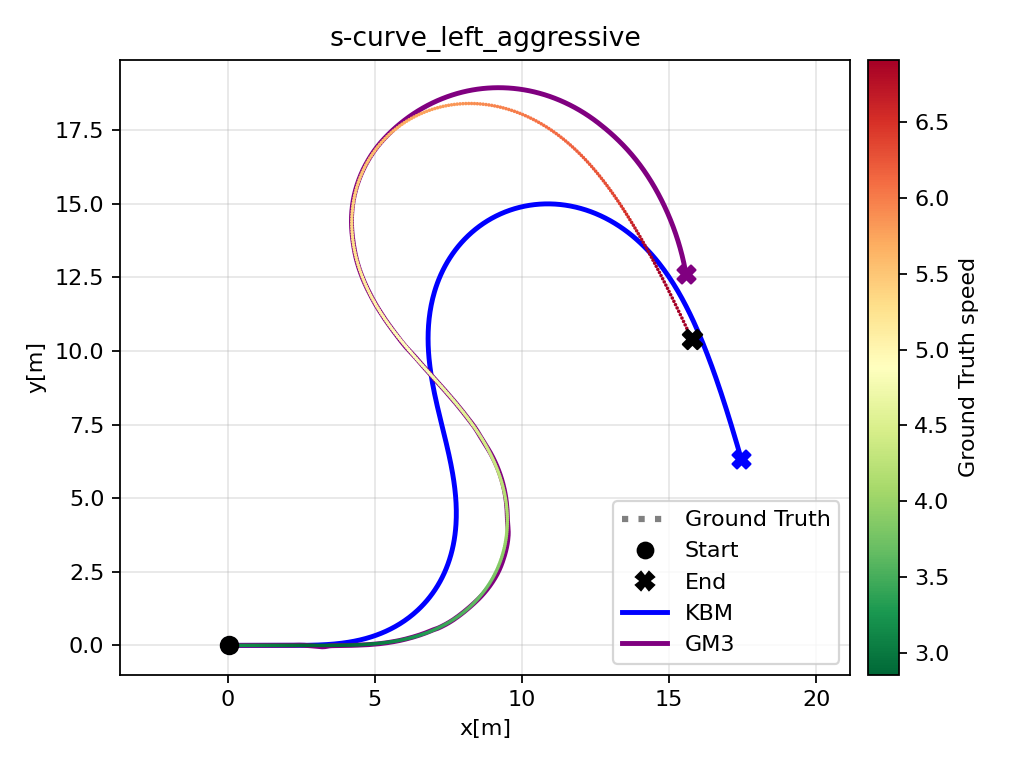}
    \caption{S-Curve}
\end{subfigure}
\hfill
\begin{subfigure}{0.32\textwidth}
    \centering
    \includegraphics[width=\linewidth]{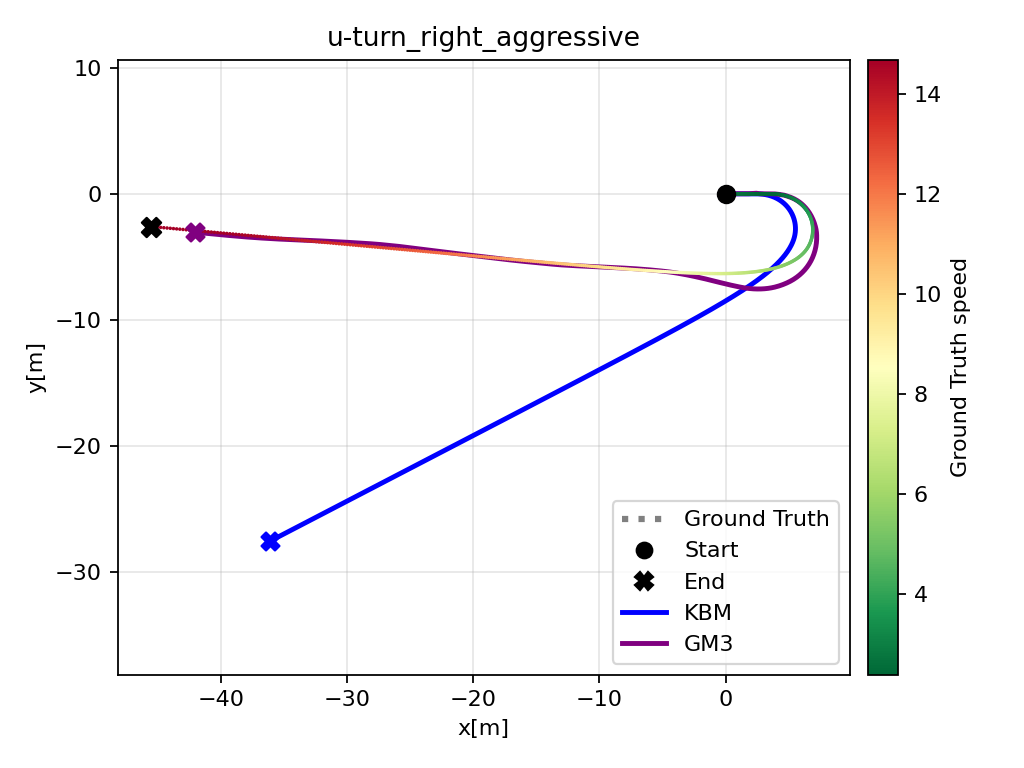}
    \caption{U-turn}
\end{subfigure}
\hfill
\begin{subfigure}{0.32\textwidth}
    \centering
    \includegraphics[width=\linewidth]{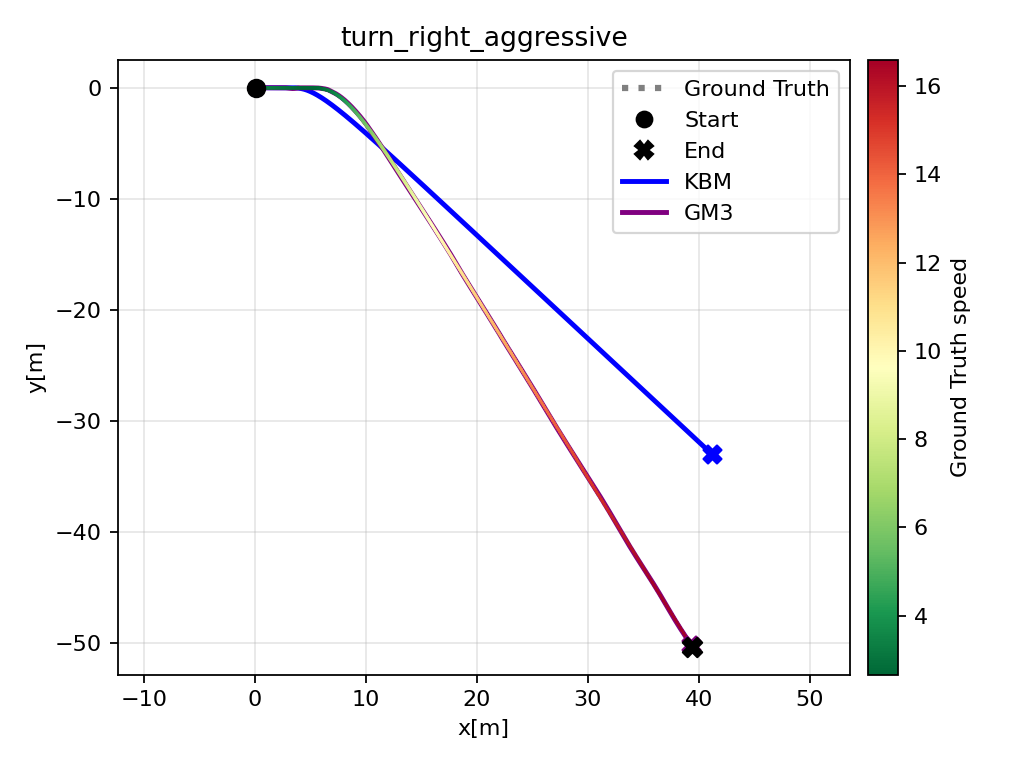}
    \caption{Right turn}
\end{subfigure}
\caption{GM3 and KBM trajectory reconstruction for S-curve, U-turn, and Right turn maneuvers, compared to the captured ground truth trajectories for these scenarios.}
\label{fig:traj_comparison}
\vspace*{-1em}
\end{figure*}

Results in table~\ref{tab:manueverresults} demonstrate that GM3 more accurately reproduces ground truth aggressive micro-mobility maneuvers than KBM. GM3 achieves lower ADE and DFD errors across the five maneuver classes with the biggest improvements observed for aggressive left and right turns. This behavior is also reflected in \ref{fig:traj_comparison}, where we see GM3 closely follows the ground-truth paths while KBM tends to overshoot or understeer in high-speed and high-curvature segments. These observations can primarily be explained by GM3's modeling of slip forces and load transfer, which allow it to better capture tire-level dynamics.

\begin{table}[t]
    \centering
    \vspace*{0.5em}
    \begin{tabular}{lccccc}
    \toprule
         & \multicolumn{2}{c}{$\downarrow$\textbf{ADE (m)}} & \phantom{a} &  \multicolumn{2}{c}{$\downarrow$ \textbf{DFD (m)}} \\
        \cmidrule{2-3} \cmidrule{5-6}
        \textbf{Mode} & \textbf{GM3} & \textbf{KBM} & \phantom{a} & \textbf{GM3} & \textbf{KBM} \\
        \midrule
        Biker (Fig~\ref{subfig:bicycle})  & \textbf{45.109} & 47.564 & \phantom{a} & 66.108 & \textbf{59.257} \\
        Skater (Fig~\ref{subfig:skateboard}) & \textbf{45.628} & 50.016 & \phantom{a} & \textbf{55.005} & 58.575 \\
        Cart (Fig~\ref{subfig:cart})  & \textbf{44.270} & 47.512 & \phantom{a} & \textbf{52.263} & 53.515 \\
        \bottomrule
    \end{tabular}
    \vspace*{-0.25em}
    \caption{\small \textbf{Mean Average Displacement Error (ADE) and discrete Fréchet distance (DFD) in meters over all trajectories for each mode from the {\em Stanford Drone Dataset DeathCircle scene}~\cite{robicquet2017learning}.} GM3 yields lower ADE than KBM across all modes, indicating a better step-by-step alignment. For DFD, GM3 improves on Skater and Cart but falls behind KBM on Biker, reflecting larger lean-induced lateral biases.}
    \label{tab:results}
    \vspace*{-1em}
\end{table}

\begin{table}[t]
    \centering
    \scriptsize
    \vspace*{0.2em}
    \begin{tabular}{lccccc}
    \toprule
         & \multicolumn{2}{c}{$\downarrow$\textbf{ADE (m)}} & \phantom{a} &  \multicolumn{2}{c}{$\downarrow$ \textbf{DFD (m)}} \\
        \cmidrule{2-3} \cmidrule{5-6}
        \textbf{Maneuver} & \textbf{GM3} & \textbf{KBM} & \phantom{a} & \textbf{GM3} & \textbf{KBM} \\
        \midrule
        S-curve & \textbf{0.7398} & 2.7537 & \phantom{a} & \textbf{4.0851} & 4.4777 \\
        Turn (Left)          & \textbf{0.2135} & 5.0892 & \phantom{a} & \textbf{0.3995} & 15.5169 \\
        Turn (Right)         & \textbf{0.2786} & 5.0892 & \phantom{a} & \textbf{0.4982} & 15.5169 \\
        U-turn (Left)        & \textbf{2.9065} & 6.7858 & \phantom{a} & \textbf{5.1678} & 26.0783 \\
        U-turn (Right)       & \textbf{2.8124} & 6.7858 & \phantom{a} & \textbf{4.8561} & 26.0783 \\
        \bottomrule
    \end{tabular}
    \vspace*{-0.25em}
    \caption{\small \textbf{Mean Average Displacement Error (ADE) and discrete Fréchet distance (DFD) across 5 aggressive maneuver classes.}}
    \label{tab:manueverresults}
    \vspace*{-2em}
\end{table}

\begin{figure}
    \centering
    \includegraphics[width=0.9\linewidth]{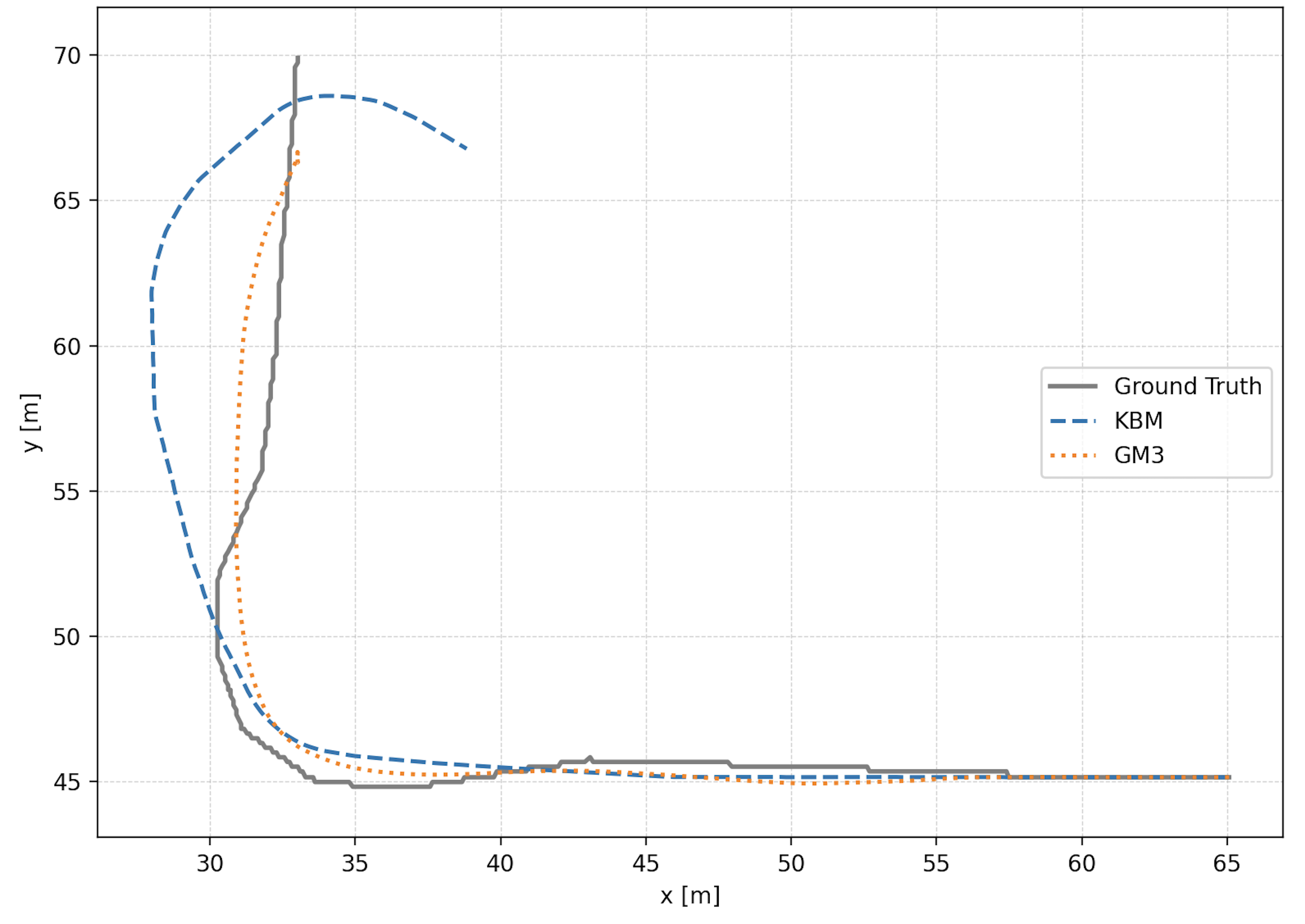}
    \vspace*{-0.5em}
    \caption{Trajectory comparison on a Biker track from the Stanford Drone Dataset DeathCircle scene. GM3 (dotted orange) follows the ground truth path (gray) more closely than the KBM (dashed blue), notably near the roundabout turn and exit, which is a \em{high curvature region}. }
    \label{fig:bikerTrajOverlay}
    \vspace*{-1em}
\end{figure}

\section{CONCLUSION}
This paper introduced the Generalized Micro-Mobility Model (GM3), a tire-level formulation built using the brush-tire model that unifies all types of micro-mobility, including common layouts such as bicycles, scooters, carts, uncommon layouts such as hoverboards and three-wheel platforms, as well as novel layouts. This model captures properties like tire slip, load transfer, and rider/vehicle lean which specialized models like Kinematic Bicycle Model (KBM) miss. Our current implementation supports explicit manual control, lean-to-steer with truck geometry, and differential drive control interfaces, but can be extended to other control methods due to GM3's modularity. To facilitate comparison, we built a model-agnostic simulation framework that decouples wheel layout from dynamics, standardizes control mappings, and supports interactive and scripted simulations. On the Stanford Drone Dataset, {\em GM3 consistently reduced average displacement errors} across bicycle, skateboard, and cart modes, while {\em achieving path-level similarity on par with the KBM} as measured by the discrete Fréchet distance. These observations highlight that {\em grounding dynamics at the tire level results in more realistic local trajectories without sacrificing global path realism}. Future work will extend the GM3 to be differentiable for learning, control, and parameter identification, collect instrumented data emphasizing lean and load transfer, study how GM3 enhances interactions between road users, and how it can be used to generate realistic data for autonomous vehicle training and personalized design of autonomous MMVs.





\section*{ACKNOWLEDGEMENTS} This work is supported in part by Dr. Barry Mersky and 
Capital One E-Nnovte Endowed Professorships, University of Maryland Distinguished University Professorship, Maryland Transportation Institute Fellowship, National 
Science Foundation, and ARL-UMD ArtIAMAS Cooperative Agreement.

\bibliographystyle{IEEEtran}
\bibliography{refs.bib}

\end{document}